\documentclass[journal]{IEEEtran}
\usepackage{amsmath,amsfonts}
\usepackage{algorithmic}
\usepackage{algorithm}
\usepackage{array}
\usepackage[caption=false,font=footnotesize,labelfont=rm,textfont=rm]{subfig}
\usepackage{textcomp}
\usepackage{stfloats}
\usepackage{url}
\usepackage{graphicx}
\usepackage{cite}
\usepackage{mathrsfs,bm}
\usepackage{amssymb}
\usepackage{mathtools}
\usepackage{amsthm}
\usepackage{multirow}
\usepackage[hidelinks]{hyperref}

\usepackage{booktabs}

\newtheorem{theorem}{Theorem}

\newtheorem{assumption}{Assumption}
\begin{document}

\title{Bilateral Sharpness-Aware Minimization for Flatter Minima}

\author{Jiaxin Deng, Junbiao Pang*, Baochang Zhang, Qingming Huang 

\IEEEcompsocitemizethanks{
\IEEEcompsocthanksitem * Corresponding author(s).
\IEEEcompsocthanksitem J. Deng and J. Pang are with the Faculty of Information Technology, Beijing University of Technology, Beijing 100124, China (email: \mbox{junbiao\_pang@bjut.edu.cn}).

\IEEEcompsocthanksitem  B. Zhang is with the School of Artificial Intelligence, BeiHang University, Beijing, 100191, China. (email: bczhang@buaa.edu.cn).

\IEEEcompsocthanksitem  Q. Huang is with the University of Chinese Academy of Sciences, Chinese Academy of Sciences (CAS), Beijing 100049, China, and the Institute of Computing Technology, CAS, Beijing
100190, China (email: qmhuang@ucas.ac.cn).
 }
}

\maketitle

\begin{abstract}

Sharpness-Aware Minimization (SAM) enhances generalization by minimizing Max-Sharpness (MaxS). Despite its practical success, we empirically found that the MaxS behind SAM’s generalization enhancements faces the ``Flatness Indicator Problem'' (FIP), where SAM only considers the flatness in the direction of gradient ascent. {This leads to high Hessian eigenvalues for the deep neural network, indicating insufficient flatness in the solution region. A better Flatness Indicator (FI) would lower these Hessian eigenvalues, resulting in a flatter minimum and improved generalization of the network.} Because SAM is inherently a greedy search method. In this paper, we propose to utilize the difference between the training loss and the minimum loss over the neighborhood surrounding the current weight, which we denote as Min-Sharpness (MinS). By merging MaxS and MinS, {we create a better FI that indicates a flatter direction during the optimization.} Specifically, we combine this FI with SAM into the proposed Bilateral SAM (BSAM) which finds {a flatter minimum} than SAM. {Under explicit smoothness, variance, update-alignment, and bounded-radius assumptions, our theoretical analysis establishes an $O(T^{-1/2})$ bound on the average expected squared gradient norm of BSAM.}
Extensive experiments demonstrate that BSAM offers superior generalization performance and robustness compared to vanilla SAM across various tasks, \textit{i.e.,} classification, transfer learning, human pose estimation, semantic segmentation, and network quantization. %

\end{abstract}


\section{Introduction}\label{sec:intro}

{Deep Neural Networks (DNNs) have achieved impressive results across various fields but are often over-parameterized, leading to overfitting and poor generalization when trained with standard loss functions~\cite{foret-2020-SAM-ICLR}.} Recent research has shown that the loss landscape is complex and non-convex with {many local minima} of different generalization abilities~\cite{li-2018-visualizing-NIPS}. Consequently, many studies have investigated the relationship between the geometry of the loss landscape and the generalization of a model~\cite{neyshabur2017exploring}\cite{jiang-2019-fantastic-ICLR}. Optimizers typically need to converge to a flat minimum~\cite{keskar-2016-large_batch-ICLR}~\cite{he2020resnet} for good generalization performance. However, the complex loss landscape of over-parameterized DNNs, with many sharp minima, makes it difficult for optimizers to find a flat minimum. 
For example, Stochastic Gradient Descent (SGD) and some of its variants serve as an implicit regularization that favors flat minima~\cite{wu2018sgd}. However, researchers prefer to use optimizers {in an explicit manner} to find the smooth loss landscapes and a flat {minimum}.

Sharpness-Aware Minimization (SAM)~\cite{foret-2020-SAM-ICLR} explicitly promotes a flat minimum by jointly minimizing the loss value and loss sharpness, leading to improved generalization performance. SAM and its variants have demonstrated state-of-the-art (SOTA) performance across various applications, such as classification~\cite{kwon-2021-asam-ICML}, transfer learning~\cite{du-2022-ESAM-ICLR}\cite{zhuang-2022-GSAM-ICLR} and domain generalization~\cite{dong2024implicit}.

However, we found that SAM faces the ``Flatness Indicator Problem'' (FIP), which results in the minimum region where the solution lies not being sufficiently flat.
{Specifically, SAM's flatness indicator considers only the sharpness in the gradient ascent direction, while ignoring the flatness along the gradient descent direction. As a result, SAM tends to converge to a locally flat minimum that is sharp on one side during training. Because SAM is inherently a greedy search method that selects the flattest direction at each update step, designing an effective Flatness Indicator (FI) to measure the flatness of the neighborhood around the current parameter point is critical. A more accurate FI can better guide the optimizer toward a flatter minimum. That is, a better FI would lead to better generalization of neural networks.}

In this paper, we propose a more effective flatness indicator to guide the search for flatter minima than SAM. We introduce Min-Sharpness (MinS), which quantifies sharpness along the gradient descent direction. Specifically, MinS is defined as the difference between the current training loss and the minimum loss within a local neighborhood in parameter space. By combining Max-Sharpness (MaxS) and MinS, we formulate a novel flatness indicator, Bilateral sharpness, and develop a new optimization method called Bilateral SAM (BSAM). {Our experiments show that BSAM finds flatter solutions and improves generalization compared with SAM.} To address potential conflicts between the MinS gradient and the descent direction of the loss, we introduce a radius decay strategy for MinS. {Under explicit smoothness, variance, update-alignment, and bounded-radius assumptions, our theoretical analysis establishes an $O(T^{-1/2})$ first-order stationarity bound for BSAM.}

In summary, our contributions are as follows:
\begin{itemize}
    \item We found that Bilateral sharpness is a {better} flatness indicator than MaxS.
    Based on this finding, we propose optimizing the bilateral sharpness of the loss landscape within the local parameter space. Our results highlight the crucial role of bilateral sharpness in achieving flatter minima for vision tasks. 
    \item We propose the BSAM method, which simultaneously optimizes the MaxS, MinS and the current training loss to find a flatter minimum. Experimental analysis of the top eigenvalues of the Hessian reveals that BSAM achieves flatter minima compared to SAM. By applying BSAM to various tasks such as classification, transfer learning, human pose estimation and semantic segmentation, BSAM significantly enhances model generalization performance. 
\end{itemize}

\section{Related Works}
\label{sec:related}
\subsection{Background of SAM}
{
Foret et al.~\cite{foret-2020-SAM-ICLR} introduced the Sharpness-Aware Minimization (SAM) method to enhance model generalization.
The optimization objective of SAM is as follows:
\begin{equation}\label{equ:sam_max}
\begin{aligned}
\mathop {\min } \limits_\mathbf{w} [(L^{max}(\mathbf{w})&- L(\mathbf{w}) ) + L(\mathbf{w}) + \lambda ||\mathbf{w}||_2^2] \\ L^{max}(\mathbf{w})&= \mathop {\max }\limits_{||\bm{\varepsilon} || \le \rho } L(\mathbf{w} + \bm{\varepsilon} ),
\end{aligned}
\end{equation}
where $\mathbf{w}$ represents the weights of the network, $\bm{\varepsilon}$ represents the perturbation of weights $\mathbf{w}$ in a Euclidean ball with the radius $\rho$ $(\rho>0)$, $L(\cdot)$ is the loss function, and $\lambda ||\mathbf{w}||_2^2$ is a standard L2 regularization term. The sharpness in SAM, referred to as Max-Sharpness (MaxS), is defined as follows:
\begin{equation}\label{equ:max_sharpness}
h^{max}(\mathbf{w}) = \mathop {\max }\limits_{||\bm{\varepsilon} || \le \rho } L(\mathbf{w} + \bm{\varepsilon}) -L(\mathbf{w}),
\end{equation}
where $h^{max}(\mathbf{w})$ quantifies the sharpness of $L(\mathbf{w})$ by measuring how quickly the training loss can be decreased by moving from $\mathbf{w}$ to nearby weight values.}

{In order to obtain $h^{max}(\mathbf{w})$, SAM utilizes Taylor expansion to search for the maximum perturbed loss ($L^{max}(\mathbf{w})$) in local parameter space as follows:
\begin{equation}\label{equ:psf}
\begin{aligned}
\mathop {\arg \max }\limits_{||\bm{\varepsilon} || \le \rho } \; L(\mathbf{w} + \bm{\varepsilon} ) & \approx \mathop {\arg \max }\limits_{||\bm{\varepsilon} || \le \rho } \; L(\mathbf{w}) + {\bm{\varepsilon} ^{\top}}{\nabla _\mathbf{w}}L(\mathbf{w}) \\
&= \mathop {\arg \max }\limits_{||\bm{\varepsilon} || \le \rho } \; {\bm{\varepsilon} ^{\top}}{\nabla _\mathbf{w}}L(\mathbf{w}).
\end{aligned}
\end{equation}
By solving Eq.~\eqref{equ:psf}, SAM obtains the perturbation as follows:
\begin{equation}\label{equ:e_max}
\begin{aligned}
\hat{ \bm{\varepsilon} } = \rho {\nabla _\mathbf{w}}L(\mathbf{w})/||{\nabla _\mathbf{w}}L(\mathbf{w})||.
\end{aligned}
\end{equation}
{This perturbation follows the gradient ascent direction and maximizes the perturbed loss ($L^{max}(\mathbf{w})$).} Substituting the perturbation $\hat{\bm{\varepsilon} }$ back into Eq.~\eqref{equ:sam_max} and differentiating, we have:
\begin{equation}\label{equ:gs}
\begin{aligned}
{\nabla _{\mathbf{w}}}&{{{L}}^{max}}({\mathbf{w}}) \approx {\nabla _{\mathbf{w}}}L({\mathbf{w}} + \hat {\bm{\varepsilon}} ({\mathbf{w}})) \\ &= \frac{{d({\mathbf{w}} + \hat {\bm{\varepsilon}} ({\mathbf{w}}))}}{{d{\mathbf{w}}}}{\nabla _{\mathbf{w}}}L({\mathbf{w}}){|_{{\mathbf{w}} + \hat {\bm{\varepsilon}} ({\mathbf{w}})}}\\
 &= {\nabla _{\mathbf{w}}}L({\mathbf{w}}){|_{{\mathbf{w}} + \hat {\bm{\varepsilon}} ({\mathbf{w}})}} + \frac{{d\hat {\bm{\varepsilon}} ({\mathbf{w}})}}{{d{\mathbf{w}}}}{\nabla _{\mathbf{w}}}L({\mathbf{w}}){|_{{\mathbf{w}} + \hat {\bm{\varepsilon}} ({\mathbf{w}})}}.
\end{aligned}
\end{equation}
By dropping the second-order terms in Eq.\eqref{equ:gs}, SAM calculates the gradient at $\mathbf{w}+\bm{\hat\varepsilon}$ as follows:
\begin{equation}\label{equ:grad_sam}
\begin{aligned}
{\nabla _{\mathbf{w}}}&{{{L}}^{max}}({\mathbf{w}}) \approx {\nabla _\mathbf{w}}L(\mathbf{w}){|_{\mathbf{w} + \bm{\hat\varepsilon} }}.
\end{aligned}
\end{equation}
Finally, SAM uses the gradients from Eq.~\eqref{equ:grad_sam} for optimization.}

\subsection{The Variants of SAM}

SAM and its variants can be roughly divided into two categories. 

{The first category focuses on enhancing the model's generalization ability.} {To address the scale-dependency problem of sharpness, where sharpness measures are sensitive to model parameter re-scaling, thereby weakening the correlation between sharpness and the generalization gap. Kwon et al. proposed {Adaptive SAM (ASAM)} ~\cite{kwon-2021-asam-ICML}, which introduces adaptive sharpness to mitigate this issue and improve generalization.}
Zhuang et al.~\cite{zhuang-2022-GSAM-ICLR} pointed out that SAM did not consistently favor the flat minima. They then proposed {surrogate Gap guided SAM (GSAM)} to improve the generalization ability by minimizing the perturbed loss and the proposed surrogate gap, which measures the difference between the maximum loss within the neighborhood and the loss at the center point.
Zhao et al. \cite{zhao-2022-penalizing-ICML} proposed a variant of SAM by penalizing the gradient norm based on the observation {that a sharper valley of loss landscape} tends to have the gradient with a larger norm. 
Zhang et al. introduced the first-order flatness \cite{zhang-2023-gradient-CVPR} which measures the maximal gradient norm within a perturbation radius of the center point. Based on this, they proposed Gradient norm Aware Minimization (GAM), {a new training approach that seeks a flatter loss landscape with uniformly small curvature, leading to improved generalization.}
{Li et al. introduced Friendly-SAM (F-SAM)} \cite{li2024friendly}, which refines SAM by removing the detrimental effects of the full gradient component of whole dataset.
In summary, these methods improve SAM from different perspectives, such as the definition of sharpness and the optimization process of sharpness, enhancing the generalization performance of the trained models. However, they both considered only unidirectional flatness.

{The second category aims to enhance the optimization efficiency of SAM.} The optimization efficiency of SAM has limited its large-scale applications since SAM needs two forward and two backward per optimization step. To address this, several methods have been proposed to accelerate SAM. For instance, 
{Zhao et al. proposed Randomized SAM (G-RST)}~\cite{zhao2022randomized}, which reduces the computational burden by randomly choosing between base optimization algorithms and sharpness-aware methods at each iteration. This approach lowers the total number of forward-backward propagation required. 
Liu et al. proposed LookSAM \cite{liu-2022-looksam-CVPR}, which accelerates SAM by reducing the frequency of perturbation computations. 
{Jiang et al. introduced Adaptive policy to Employ SAM (AE-SAM) }\cite{jiang-2022-aesam-ICLR} to speed up SAM by adaptively choosing when SAM is used based on the loss landscape.
{Ni et al. proposed K-SAM }\cite{ni2022k}, which reduces SAM's computational cost by computing gradients only with the top-$k$ samples with the highest loss. 
{Du et al. proposed Efficient SAM (ESAM)} \cite{du-2022-ESAM-ICLR} to accelerate SAM by two strategies: Stochastic Weight Perturbation, which approximates sharpness by perturbing a subset of weights, and Sharpness-Sensitive Data Selection, which uses a carefully chosen data subset to optimize the loss. 
Becker et al. proposed Momentum-SAM (MSAM) \cite{becker2024momentum} without requiring additional forward and backward computations by using the momentum direction as an approximation of the perturbation direction. 
In summary, these methods greatly improve the computational efficiency of SAM-based methods, but most of them come at the cost of sacrificing model generalization performance.

\begin{figure*}[!t]
\centering
\subfloat[The MaxS, MinS and BilS during training.]{\includegraphics[width=2.3in]{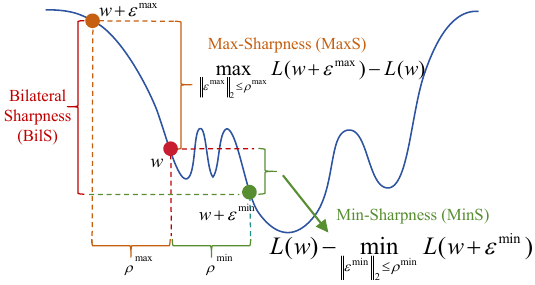}
\label{fig:min_sharpness1}}
\hfil
\subfloat[Loss landscape after training with SAM.]{\includegraphics[width=2.0in]{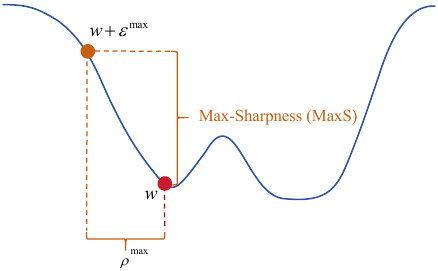}
\label{fig:min_sharpness2}}
\hfil
\subfloat[Loss landscape after training with BSAM.]{\includegraphics[width=2.0in]{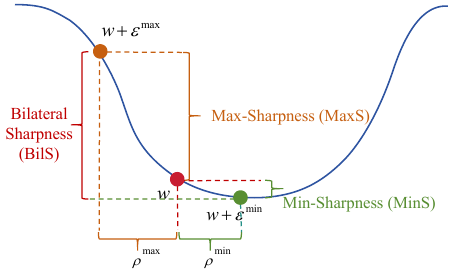}
\label{fig:min_sharpness3}}
\caption{Illustration of the notations of the MaxS, MinS and BilS.}
\label{fig:min_sharpness}
\end{figure*}

\section{Method}

\subsection{Min-Sharpness}

Max-Sharpness (MaxS) in Eq. \eqref{equ:max_sharpness} just captures the difference between the maximum loss of perturbed points and the loss at the current parameter point, promoting flatness from a single direction.

{To describe flatness along the gradient descent direction, we define Min-Sharpness (MinS) as follows:
\begin{equation}\label{equ:sam-min}
\begin{aligned}
h^{min}(\mathbf{w}) = L(\mathbf{w}) - \mathop {\min }\limits_{||\bm{\varepsilon}^{min} || \le \rho^{min} } L(\mathbf{w} + \bm{\varepsilon}^{min} ),
\end{aligned}
\end{equation}
}
where $\bm{\varepsilon}^{min}$ represents weight perturbations in Euclidean ball with radius $\rho^{min}$. MinS measures how quickly the training loss decreases as the parameters move from $\mathbf{w}$ to a nearby value.

{We obtain the optimal ${{\bm{\varepsilon}} ^{min}}$ that minimizes the loss $ L(\mathbf{w} + \bm{\varepsilon}^{min} )$ in Eq.~\eqref{equ:sam-min} via a first-order Taylor expansion of $L(\mathbf{w}+\bm{\varepsilon}^{\min})$
around 0 as follows:}
\begin{equation}\label{equ:taylor_min}
\begin{aligned}
{\hat{\bm{\varepsilon}} ^{min}} & = \mathop {\arg \min }\limits_{{{\left\| {{\bm{\varepsilon} ^{min }}} \right\|}_p} \le \rho^{min}} L(\mathbf{w} + {\bm{\varepsilon} ^{min }}) \\ & \approx \mathop {\arg \min }\limits_{{{\left\| {{\bm{\varepsilon} ^{min }}} \right\|}_p} \le \rho^{min}} L(\mathbf{w}) + {\bm{\varepsilon} ^{{{min }^{\top}}}}{\nabla _\mathbf{w}}L(\mathbf{w})\\ & \approx \mathop {\arg \min }\limits_{{{\left\| {{\bm{\varepsilon} ^{min }}} \right\|}_p} \le \rho^{min}} {\bm{\varepsilon} ^{{{min }^{\top}}}}{\nabla _\mathbf{w}}L(\mathbf{w}).
\end{aligned}
\end{equation}

The proof in the Appendix A shows that $L(\mathbf{w} + \bm{\varepsilon}^{min} )$ reaches its minimum value when \\ 
\begin{equation}\label{equ:hat_min}
\begin{aligned}
\hat{ \bm{\varepsilon}}^{min}  = -\rho^{min}\frac{ {\nabla _\mathbf{w}}L(\mathbf{w})}{||{\nabla _\mathbf{w}}L(\mathbf{w})||},
\end{aligned}
\end{equation}
where $\nabla _\mathbf{w}L(\mathbf{w})$ is the gradient at $\mathbf{w}$, and $||\cdot||$ represents the L2-norm. We observe that the direction of $\hat{ \bm{\varepsilon}}^{min}$ is along the gradient descent direction, which is opposite to the perturbation direction in vanilla SAM. 

\subsection{BSAM}\label{sec:bsam}

{We propose a method to jointly optimize the training loss ($L(\mathbf{w})$), MaxS (Eq.~\ref{equ:max_sharpness}) and MinS (Eq.~\ref{equ:sam-min}) to find flatter minima. The overall loss function can be written as follows:}
\begin{equation}\label{equ:loss_bsam}
\begin{aligned}
&\hat{L}(\mathbf{w}) = \underbrace {L(\mathbf{w})}_{\text{Training loss}} \\
&+ \underbrace {\left[ {\mathop {\max }\limits_{\left\| {{\bm{\varepsilon} ^{\max }}} \right\| \le {\rho ^{\max }}} {{L}}(\mathbf{w} + {\bm{\varepsilon} ^{\max }}) - \mathop {\min }\limits_{\left\| {{\bm{\varepsilon} ^{\min }}} \right\| \le {\rho ^{\min }}} {{L}}(\mathbf{w} + {\bm{\varepsilon} ^{\min }})} \right]}_{\text{Bilateral Sharpness (BilS)}}.
\end{aligned}
\end{equation}
Eq. \eqref{equ:loss_bsam} ensures that, during training, the loss not only decreases but also becomes more balanced on both sides of the current parameter point, resulting in a flatter parameter space around it.

\textbf{Bilateral Sharpness (BilS) is a better flatness indicator than MaxS:} 
Fig. \ref{fig:min_sharpness1} illustrates the concepts of the MaxS, MinS and BilS. MaxS only describes the flatness of the loss landscape on one side of the current parameter point.
Due to the greedy nature of SAM, it seeks a flat region between the maximum loss point of the current optimization step and the current point. 
Consequently, the loss landscape may still be insufficiently flat after training, as illustrated in Fig. \ref{fig:min_sharpness2}. 
In contrast, the proposed BilS is a new flatness indicator that identifies a flatter direction than SAM, making the parameter update direction more accurate and stable, as a comparison between Fig. \ref{fig:min_sharpness1} and Fig. \ref{fig:min_sharpness3}.

{Similar to Eq.~\eqref{equ:gs}, the gradient of MinS at the perturbation point $\mathbf{w} + \bm{\varepsilon}^{\min}$ can be written as follows:
\begin{equation}\label{equ:gradient_bsam}
\begin{aligned}
&{\nabla _{\mathbf{w}}}{{{L}}^{min}}({\mathbf{w}})
\approx {\nabla _\mathbf{w}}L(\mathbf{w} + {{\hat{\bm{\varepsilon}}}^{\min }}(\mathbf{w})) \\
 &= {\nabla _\mathbf{w}}L(\mathbf{w}){|_{\mathbf{w} + {{\hat {\bm{\varepsilon}}}^{\min }}(\mathbf{w})}} + \frac{{d{{\hat {\bm{\varepsilon}}}^{\min }}(\mathbf{w})}}{{d\mathbf{w}}}{\nabla _\mathbf{w}}L(\mathbf{w}){|_{\mathbf{w} + {{\hat {\bm{\varepsilon}}}^{\min }}(\mathbf{w})}}.
\end{aligned}
\end{equation}}
To further accelerate computation, we also drop the second-order terms. The gradient of MinS is approximated as follows:
\begin{equation}\label{equ:grad_sam_min}
\begin{aligned}
{\nabla _{\mathbf{w}}}{{{L}}^{min}}({\mathbf{w}}) \approx {\nabla _\mathbf{w}}L(\mathbf{w}){|_{\mathbf{w} + \bm{\hat\varepsilon}^{min} }}.
\end{aligned}
\end{equation}

{By combining the gradient of the training loss (${\nabla _\mathbf{w}}L(\mathbf{w})$) with Eq.~\eqref{equ:grad_sam} and Eq.~\eqref{equ:grad_sam_min}, the gradient for updating parameters in BSAM is as follows:}
\begin{equation}\label{equ:grad_L}
\begin{aligned}
&{\nabla _\mathbf{w}}\hat L(\mathbf{w}) = \underbrace {{\nabla _\mathbf{w}}L(\mathbf{w})}_{{\rm{Gradient\;of\;the\;training\;loss}}}\\
 &\quad+ \underbrace {({\nabla _\mathbf{w}}L(\mathbf{w}){|_{\mathbf{w} + {{\hat{ \bm{\varepsilon}} }^{\max }}(\mathbf{w})}} - {\nabla _\mathbf{w}}L(\mathbf{w}){|_{\mathbf{w} + {{\hat{ \bm{\varepsilon} }}^{\min }}(\mathbf{w})}})}_{{\rm{Gradients\;that\;promote\;flat\;minima}}}.
\end{aligned}
\end{equation}
BSAM employs the base optimizer to perform gradient descent using the gradient from Eq.~\eqref{equ:grad_L}. The first term in this gradient promotes finding regions of local minima, while the second term encourages locating flatter regions. 

\begin{figure*}[t!]
\centering
\subfloat[Gradient conflict occurs when $\rho^{min}$ is too large.]{\includegraphics[width=1.8in]{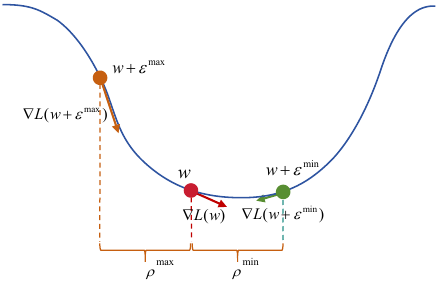}%
\label{fig:gc1}}
\hfil
\subfloat[There is no gradient conflict when $\rho^{min}$ is small.]{\includegraphics[width=1.8in]{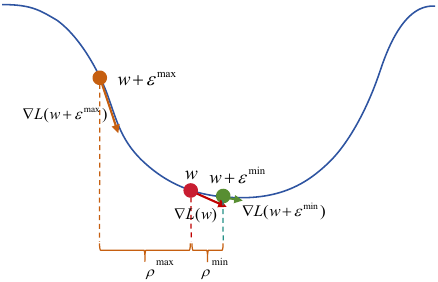}%
\label{fig:gc2}}
\hfil
\subfloat[$\rho^{min}$ changes with the learning rate.]{\includegraphics[width=1.5in]{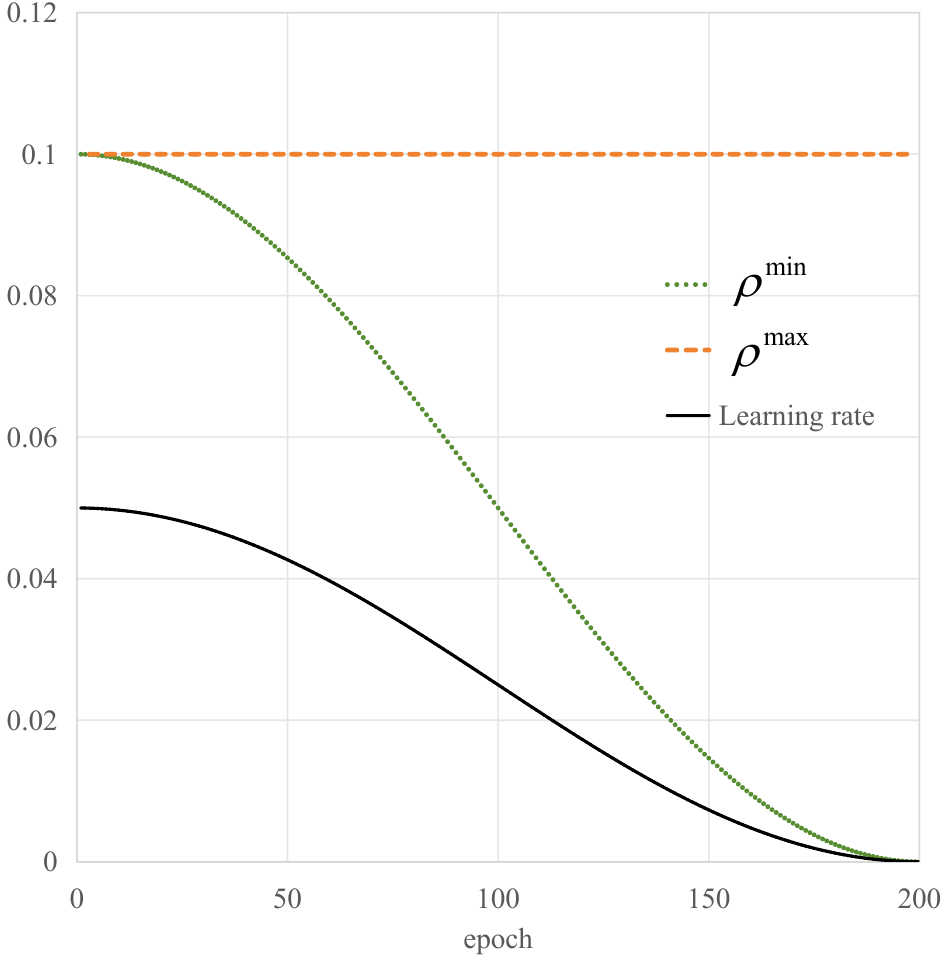}%
\label{fig:lr}}
\caption{The occurrence of gradient conflict under different $\rho^{min}$ and the variation of $\rho^{min}$ with learning rate in BSAM.}
\label{fig:gc}
\end{figure*}

\begin{figure*}[t!]
\centering
\subfloat[Fixed $\rho^{min}$.]{\includegraphics[width=1.7in]{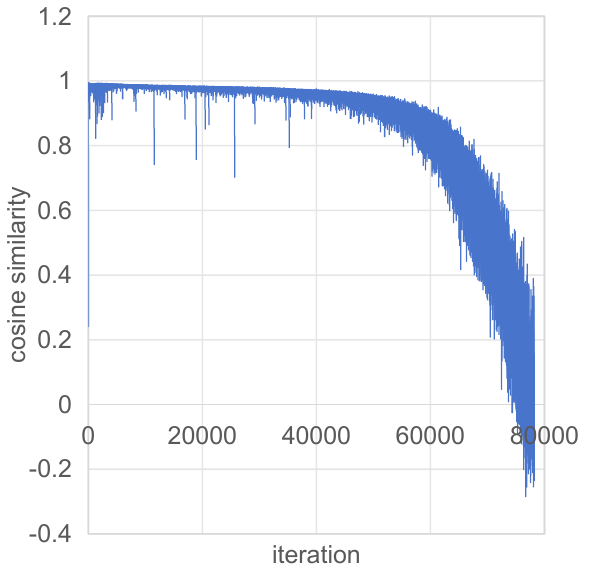}%
\label{fig:cos_sim1}}
\hfil
\subfloat[$\rho^{min}$ decreases from 0.1 to 0.05.]{\includegraphics[width=1.7in]{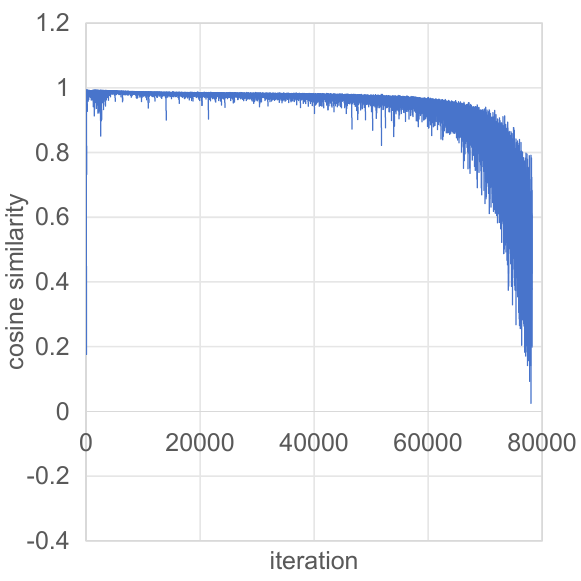}%
\label{fig:cos_sim2}}
\hfil
\subfloat[$\rho^{min}$ decreases from 0.1 to 0.]{\includegraphics[width=1.7in]{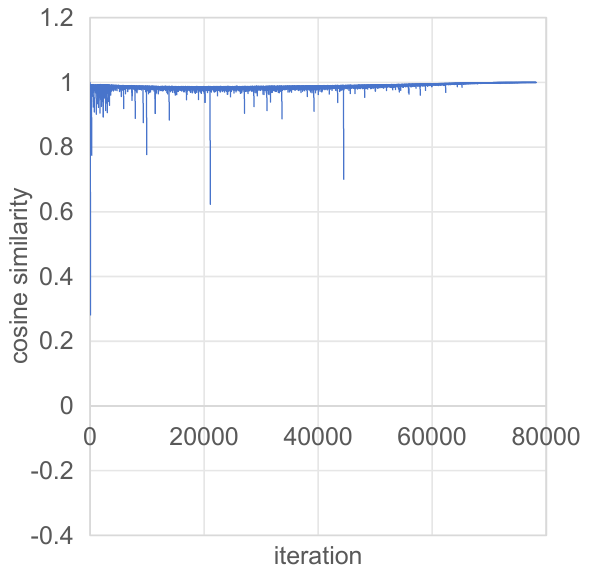}%
\label{fig:cos_sim3}}
\caption{The cosine similarity between the gradients at the point 
$\mathbf{w}$ and the point $\mathbf{w} + {\hat{\bm{\varepsilon}} }^{\min }$ in training stage under the different settings of ${\rho}^{min}$.}
\label{fig:cos_sim}
\end{figure*}

\textbf{Solve Gradient Conflict Problem:} However, in the later stages of training, as illustrated in Fig. \ref{fig:gc1}, a large fixed $\rho^{min}$ can cause the perturbation point $\mathbf{w}^{min}$ ($\mathbf{w}^{min}=\mathbf{w} + {\hat{\bm{\varepsilon}} }^{\min }$) to overshoot the minimum point. The directions of ${{\nabla_\mathbf{w}}L(\mathbf{w})}$ and ${{\nabla_\mathbf{w}}L(\mathbf{w}+{\hat{\bm{\varepsilon}}}^{\min })}$ are opposite, causing the Gradient Conflict Problem (GCP) between the gradient of the original point $\mathbf{w}$ and the gradient of the perturbed point $\mathbf{w}^{min}$. {Fig.~\ref{fig:cos_sim} shows the cosine similarity between the gradients at $\mathbf{w}$ and $\mathbf{w} + {\hat{\bm{\varepsilon}} }^{\min }$ during ResNet-18 training with BSAM on CIFAR-100 dataset. In Fig.~\ref{fig:cos_sim1}, with fixed $\rho^{\min}$, the cosine similarity is high in the early stages of training (\textit{e.g.}, before 60000 iterations), indicating no gradient conflict. However, in the later stages of training (\textit{e.g.}, after 70000 iterations), the cosine similarity decreases rapidly and may even drop below zero, indicating that the gradient conflict occurs.} 

{Although gradient decomposition could address the GCP, we chose not to use it due to the additional computational overhead it introduces, which is undesirable in practical implementations.} {Instead, we observed that gradually decreasing $\rho^{\min}$ helps reduce gradient conflicts during training. We assume that the gradient within the local range of the parameter space at point $\mathbf{w}$ does not change abruptly. {When $\rho^{\min}$ is small, the gradient at the perturbation point is close to that at $\mathbf{w}$, reducing the likelihood of gradient conflict.} Fig. \ref{fig:gc2} illustrates that after decreasing $\rho^{min}$ during training, there is no gradient conflict between ${{\nabla_\mathbf{w}}L(\mathbf{w})}$ and ${{\nabla_\mathbf{w}}L(\mathbf{w}+{\hat{\bm{\varepsilon}}}^{\min })}$.  Based on this, we designed a scheduler that dynamically adjusts $\rho^{\min}$, decreasing it as the learning rate decreases. The scheduling process is defined as follows:}
\begin{equation}\label{equ:rho_change}
\begin{aligned}
\rho _{t}^{\min}=\check{\rho}^{\min}+\frac{\left( \hat{\rho}^{\min}-\check{\rho}^{\min} \right) \left( lr_t-lr^{\min} \right)}{lr^{\max}-lr^{\min}},
\end{aligned}
\end{equation}
where $lr_t$ is the learning rate at $t$-th iteration, $lr^{\max}$ and $lr^{\min}$ are the maximum and minimum learning rates, respectively. $\rho _{t}^{\min}$ is the value of $\rho^{\min}$ at $t$-th iteration, $\hat{\rho}^{min}$ and $\check{\rho}^{min}$ are hyper-parameters, representing the maximum and minimum value of $\rho^{min}_t$, respectively. In our experimental setup described in Section \ref{classification_setup}, the changes in $\rho^{min}_t$ and learning rate during training are shown in Fig.~\ref{fig:lr}. {Comparing Fig.~\ref{fig:cos_sim2} and Fig. \ref{fig:cos_sim3}, the latter shows that the cosine similarity between gradients at $\mathbf{w}$ and $\mathbf{w}^{\min}$ remains above 0, indicating no gradient conflict.} 

{For a nonzero gradient, the normalized perturbation in Eq.~\eqref{equ:hat_min} has norm $\rho_t^{min}$, regardless of the gradient magnitude. Under mini-batch loss smoothness, reducing $\rho_t^{min}$ bounds the difference between the gradients at the original and minimum-side perturbed points by $\tau\rho_t^{min}$. This motivates the radius decay strategy, although radius decay alone does not guarantee alignment of the complete BSAM update with the full training gradient.}

\textbf{Balance between the MaxS, MinS, and training loss:} For convenience, we denote the perturbation ${\hat{\bm{\varepsilon}}}$ {in Eq.~\eqref{equ:e_max}} as ${\hat{\bm{\varepsilon}}}^{max}$. Our experiments reveal a disparity in the gradient magnitudes between the maximum perturbation parameter point $\mathbf{w}^{max}$ ($\mathbf{w}^{max}=\mathbf{w} + {\hat{\bm{\varepsilon}} }^{max }$) and the minimum perturbation parameter point $\mathbf{w}^{min}$ ($\mathbf{w}^{min}=\mathbf{w} + {\hat{\bm{\varepsilon}} }^{min}$) during training. Generally, the gradient magnitude at the maximum perturbation point is larger than that at the minimum point, which may diminish the effectiveness of MinS. 

\begingroup
To balance these magnitudes, we scale the gradient at the minimum point ${\nabla _\mathbf{w}}L(\mathbf{w}^{min})$
to match the gradient magnitude at the maximum point ${\nabla _\mathbf{w}}L(\mathbf{w}^{max})$, while retaining the training-loss gradient. The resulting BSAM update direction is defined as follows:
\begin{equation}\label{equ:grad_L_s}
\begin{aligned}
\mathbf{d}(\mathbf{w}) &= {{\nabla _\mathbf{w}}L(\mathbf{w})} + {\nabla _\mathbf{w}}L(\mathbf{w}){|_{\mathbf{w} + {{\hat{ \bm{\varepsilon}} }^{max }}(\mathbf{w})}} \\
 &- \frac{||{\nabla _\mathbf{w}}L(\mathbf{w}){|_{\mathbf{w} + {{\hat{ \bm{\varepsilon}} }^{max }}(\mathbf{w})}}||}{||{\nabla _\mathbf{w}}L(\mathbf{w}){|_{\mathbf{w} + {{\hat{ \bm{\varepsilon}} }^{min }}(\mathbf{w})}}||}\cdot {\nabla _\mathbf{w}}L(\mathbf{w}){|_{\mathbf{w} + {{\hat{ \bm{\varepsilon}} }^{min }}(\mathbf{w})}}.
\end{aligned}
\end{equation}
The scaling factor balances the two perturbed gradient norms; the resulting direction need not be the exact gradient of the objective in Eq.~\eqref{equ:loss_bsam}. Algorithm \ref{algorithm:1} applies this construction to the same mini-batch at all three parameter points.
\par\endgroup

\begin{algorithm}[t]
    \caption{{Pseudocode of the proposed method}}
    {\bf Require:}
    {The training dataset, initial weights $\mathbf{w}_0$, iteration budget $T$, learning rates $\eta_t$, the batch size $b$, parameters $\rho^{max}$, $\hat{\rho}^{min}$ and $\check{\rho}^{min}$.}
    \begin{algorithmic}[1]
    \FOR{{$t = 0,1,\ldots,T-1$}}
    \STATE Randomly sample a mini-batch $\mathbf{B}$ of size $b$;
    \STATE $\mathbf{g}_t=\nabla _\mathbf{w} L_\mathbf{B}(\mathbf{w}_t)$;
    \STATE {Compute perturbation $\hat{\bm{\varepsilon}}_t^{max} = \rho^{max}\mathbf{g}_t/\|\mathbf{g}_t\|_2$;}
    \STATE Calculate the gradient at $\mathbf{w}_t+\hat{ \bm{\varepsilon}}_t^{max}$: \\ \quad $\mathbf{g}_t^{max}=\nabla _\mathbf{w} L_\mathbf{B}(\mathbf{w}_t+\hat{ \bm{\varepsilon}}_t^{max})$;
    \STATE {Update $\rho_t^{min}$ using Eq.~\eqref{equ:rho_change} with $lr_t=\eta_t$;}
    \STATE {Compute perturbation $\hat{\bm{\varepsilon}}_t^{min} = -\rho_t^{min}\mathbf{g}_t/\|\mathbf{g}_t\|_2$;}
    \STATE Calculate the gradient at $\mathbf{w}_t+\hat{ \bm{\varepsilon}}^{min}_t$: \\ \quad $\mathbf{g}^{min}_t=\nabla _{\mathbf{w}} L_\mathbf{B}(\mathbf{w}_t+\hat{ \bm{\varepsilon}}^{min}_t)$; 
    \STATE Update the weights by: \\ \quad {${\mathbf{w}_{t+1}} = {\mathbf{w}_{t}} - \eta_t(\mathbf{g}_{t} + \mathbf{g}^{max}_{t}-\frac{||\mathbf{g}^{max}_{t}||}{||\mathbf{g}^{min}_{t}||}\mathbf{g}^{min}_{t})$;}
    \ENDFOR
    \end{algorithmic}
    \label{algorithm:1}
\end{algorithm}

\textbf{Time complexity:} 
{When the model has $N$ weights and the batch size of optimizer is $b$, the time complexity of SGD is $O(bN)$. Assuming the majority of the time spent by the optimizer is on the forward and backward, SAM requires two forward and backward per optimization step, giving a complexity of $O(2bN)$. BSAM requires three forward and backward per optimization step, resulting in $O(3bN)$, making it approximately 1.5 times more computationally expensive than SAM.} {This increase in complexity is independent of dataset size, as it only impacts the forward and backward propagation steps, without requiring data reloading or additional data-dependent operations.} However, in some scenarios, sacrificing efficiency is necessary to achieve better performance. Addressing this trade-off remains an important direction for future research.

\subsection{An example of the training characteristics of BSAM} %
\begin{figure}[!t]
\centering
\subfloat{\includegraphics[width=1.8in]{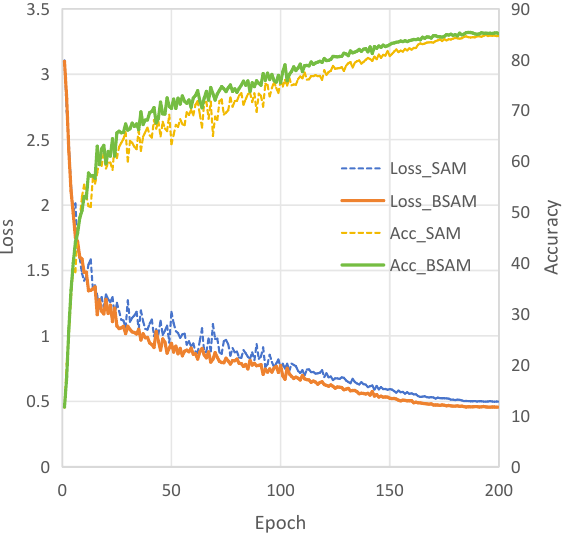}}
\caption{The loss curves and accuracy ones of SAM and BSAM when WideResNet-28-10 was used.}
\label{fig:loss_acc}
\end{figure}

Fig. \ref{fig:loss_acc} illustrates the loss and accuracy curves for both SAM and BSAM on the CIFAR-100 dataset. {Two key observations can be made:}
\begin{itemize}
    \item \textbf{The loss of BSAM decreases faster than that of SAM, while the accuracy of BSAM improves more rapidly than that of SAM.} This indicates that the MinS guides the parameter update direction more accurately at the early stages of training. Additionally, optimizing the MaxS and MinS with non-conflicting gradients accelerates the gradient descent process.
    \item \textbf{The loss curve of BSAM shows less fluctuation compared to SAM from epoch 0 to around 150.} This behavior is explained by the fact that BSAM's trajectory follows {a flatter direction than SAM's.} It indicates that BSAM searches for flatter regions than SAM during the optimization process. 
\end{itemize}

\subsection{Convergence analysis}\label{sec:convergence}
\begingroup
We analyze the first-order convergence of the BSAM update with an SGD base optimizer, without momentum or adaptive preconditioning. Let $L=L_{\mathbf{D}}$ be the average loss over the full training set and $\mathbf{g}_t=\nabla L(\mathbf{w}_t)$ its gradient. At each iteration, a mini-batch $\mathbf{B}_t$ of fixed size is sampled uniformly with replacement, independently of the previous training history. Let $\mathcal{F}_t$ denote the history before this draw, and write $\mathbb{E}_t[\cdot]=\mathbb{E}[\cdot\mid\mathcal{F}_t]$. In this section, we denote the mini-batch gradients by
\begin{equation}\label{equ:theory_gradients}
\begin{aligned}
\mathbf{h}_t &= \nabla L_{\mathbf{B}_t}(\mathbf{w}_t),\\
\hat{\mathbf{h}}_t &= \nabla L_{\mathbf{B}_t}(\mathbf{w}_t+\hat{\bm{\varepsilon}}_t^{max}),\\
\check{\mathbf{h}}_t &= \nabla L_{\mathbf{B}_t}(\mathbf{w}_t+\hat{\bm{\varepsilon}}_t^{min}),
\end{aligned}
\end{equation}
where $\hat{\bm{\varepsilon}}_t^{max}=\rho_t^{max}\mathbf{h}_t/\|\mathbf{h}_t\|_2$ and $\hat{\bm{\varepsilon}}_t^{min}=-\rho_t^{min}\mathbf{h}_t/\|\mathbf{h}_t\|_2$. Define
\begin{equation}\label{equ:theory_direction}
\begin{aligned}
\varsigma_t&=\frac{\|\hat{\mathbf{h}}_t\|_2}{\|\check{\mathbf{h}}_t\|_2},&
\mathbf{c}_t&=\hat{\mathbf{h}}_t-\varsigma_t\check{\mathbf{h}}_t,\\
\mathbf{d}_t&=\mathbf{h}_t+\mathbf{c}_t,&
\mathbf{w}_{t+1}&=\mathbf{w}_t-\eta_t\mathbf{d}_t.
\end{aligned}
\end{equation}
We require $\|\mathbf{h}_t\|_2>0$ and $\|\check{\mathbf{h}}_t\|_2>0$ almost surely, so that these normalizations are well defined. The ratio $\varsigma_t$ is random and may vary across iterations.
\par\endgroup

\begingroup
\begin{assumption}[Smoothness]\label{assumpion:smooth}
The full training loss and every possible mini-batch loss are $\tau$-Lipschitz smooth, with $\tau>0$. For all $\mathbf{w},\mathbf{v}$ and mini-batches $\mathbf{B}$,
\begin{equation}\label{equ:theory_smoothness}
\begin{aligned}
\|\nabla L(\mathbf{w})-\nabla L(\mathbf{v})\|_2
&\le\tau\|\mathbf{w}-\mathbf{v}\|_2,\\
\|\nabla L_{\mathbf{B}}(\mathbf{w})-\nabla L_{\mathbf{B}}(\mathbf{v})\|_2
&\le\tau\|\mathbf{w}-\mathbf{v}\|_2.
\end{aligned}
\end{equation}
\end{assumption}
\begin{assumption}[Unbiasedness and bounded variance]\label{assumpion:bound_var}
There is a constant $\sigma^2\ge0$, independent of $t$ and $T$, such that
\begin{equation}\label{equ:theory_variance}
\mathbb{E}_t[\mathbf{h}_t]=\mathbf{g}_t,\qquad
\mathbb{E}_t\|\mathbf{h}_t-\mathbf{g}_t\|_2^2\le\sigma^2.
\end{equation}
\end{assumption}
Smoothness and bounded stochastic-gradient variance are standard conditions in non-convex optimization~\cite{zhuang-2022-GSAM-ICLR,li2024friendly}. To control the effect of the bilateral correction, we additionally impose the following conditions.

\begin{assumption}[Update alignment]\label{assumption:alignment}
There is a constant $a>0$, independent of $t$ and $T$, such that almost surely
\begin{equation}\label{equ:theory_alignment}
\mathbb{E}_t[\mathbf{g}_t^\top\mathbf{d}_t]\ge a\|\mathbf{g}_t\|_2^2.
\end{equation}
\end{assumption}
By Assumption~\ref{assumpion:bound_var}, this is equivalent to $\mathbb{E}_t[\mathbf{g}_t^\top\mathbf{c}_t]\ge-(1-a)\|\mathbf{g}_t\|_2^2$. It limits how much the bilateral correction can oppose the full training gradient. This is an additional condition on the update direction and does not follow from smoothness and bounded variance alone.

\begin{assumption}[Bounded perturbation radii]\label{assumption:radii}
The nonnegative radii satisfy $\rho_t^{max}+\rho_t^{min}\le R$ almost surely, for a finite constant $R\ge0$ independent of $t$ and $T$.
\end{assumption}

\begin{theorem}[First-order convergence of BSAM]\label{theorem1}
Suppose Assumptions~\ref{assumpion:smooth}--\ref{assumption:radii} hold, $\mathbf{w}_0$ is deterministic, and $L(\mathbf{w})\ge L_{\inf}>-\infty$. For a given iteration budget $T\ge\max\{1,4\tau^2/a^2\}$, use the fixed learning rate $\eta_t=\eta=1/\sqrt{T}$. Then the updates in Eq.~\eqref{equ:theory_direction} satisfy
\begin{equation}\label{equ:bsam_convergence}
\begin{aligned}
\frac{1}{T}\sum_{t=0}^{T-1}\mathbb{E}\|\mathbf{g}_t\|_2^2
&\le \frac{2\bigl(L(\mathbf{w}_0)-\mathbb{E}[L(\mathbf{w}_T)]\bigr)}{a\sqrt{T}}\\
&\quad+\frac{2\tau(\sigma^2+4\tau^2R^2)}{a\sqrt{T}}.
\end{aligned}
\end{equation}
The expectations in this bound are over all mini-batch draws.
\end{theorem}
The full proof is given in Appendix~\ref{app:convergence}. Its key step bounds the complete correction $\|\mathbf{c}_t\|_2\le2\tau R$, rather than imposing an upper bound on the ratio $\varsigma_t$.

\textbf{Remark:} Since $L$ is bounded below and $a,\tau,\sigma,R$ are independent of $T$, Eq.~\eqref{equ:bsam_convergence} gives an $O(T^{-1/2})$ bound on the average expected squared gradient norm. Equivalently, an iterate selected uniformly from $\{\mathbf{w}_0,\ldots,\mathbf{w}_{T-1}\}$ has an expected squared gradient norm of this order. This is a first-order stationarity guarantee; it does not establish convergence of the entire parameter sequence to a local minimum or a comparison of Hessian eigenvalues with SAM. The ratio $\varsigma_t$ may exceed one and need not be constant. Bounded radii control the correction magnitude, while Assumption~\ref{assumption:alignment} separately controls its alignment. The fixed learning rate in this theorem is chosen for each horizon $T$; the theorem does not establish a rate for the decaying learning-rate schedules used in our experiments.
\par\endgroup

\section{Experimental Results}

\subsection{Parameter Studies}

The value of $\rho^{min}$ affects whether a parameter point remains a minimum within a local region of $\mathbf{w}$ after perturbation of the gradient descent direction, which impacts the gradient direction  and model performance. We study the effect of $\rho^{min}$ on accuracy using Resnet-18 on the CIFAR-10 and CIFAR-100 datasets. The results are summarized in Tab. \ref{result_rho_min}. For CIFAR-10, when $\rho^{min}$ is very small, the accuracy is close to that of SAM (as shown in Tab. \ref{result_cifar}), because the gradients at the current parameter point and the perturbed point in the gradient descent direction are nearly identical, minimizing the impact of MinS. {When $\rho^{min}$ is large (\textit{e.g.} $\rho^{min}=0.1$ or $0.25$), gradient conflicts as described in Section \ref{sec:bsam} may occur, reducing accuracy and performing worse than SAM. The best performance is achieved with $\rho^{min}=0.05$.} {For CIFAR-100, as the data complexity increases,} using the values of rho listed in the table for training the model consistently yields better accuracy than SAM. In summary, comparisons show that $\rho^{min}$ is less sensitive in more complex scenarios.

\begin{table}[]
\centering
\begin{tabular}{lcc}
\hline
$\rho^{min}$ & CIFAR-10 & CIFAR-100 \\ \hline
0.01                &96.75\scriptsize{$\pm$0.09}          & 81.56\scriptsize{$\pm$0.05}\\
0.025               &96.78\scriptsize{$\pm$0.02}          &81.29\scriptsize{$\pm$0.11}           \\
0.05                &96.82\scriptsize{$\pm$0.12}          &81.65\scriptsize{$\pm$0.04}           \\
0.1                 &96.66\scriptsize{$\pm$0.13}          &81.48\scriptsize{$\pm$0.18}           \\
0.25                &96.18\scriptsize{$\pm$0.09}          &81.58\scriptsize{$\pm$0.25}           \\ \hline
\end{tabular}
\caption{Test accuracy (\%) with different $\rho^{min}$ on CIFAR-10 and CIFAR-100.}
\label{result_rho_min}
\end{table}

\subsection{Image Classification}
\subsubsection{Setup}\label{classification_setup}
{To evaluate BSAM's effectiveness, we conduct experiments on the CIFAR-10 and CIFAR-100 \cite{krizhevsky2009learning} datasets across a variety of architectures}, i.e. ResNet-18 \cite{he2016deep}, WideResNet-28-10 \cite{zagoruyko2016wide} and PyramidNet-110 \cite{han2017deep}. These three models are trained for 200 epochs. For ResNet-18 and WideResNet-28-10, the initial learning rate is set to 0.05 with a cosine schedule, the momentum and weight decay are set to 0.9 and 0.001. For PyramidNet-110, the initial learning rate is 0.1, the momentum and weight decay are set to 0.9 and 0.0005. For CIFAR-10, $\rho^{max}$ and $\hat{\rho}^{min}$ are set to 0.05, and for CIFAR-100, they are set to 0.1. $\check{\rho}^{min}$ is set to 0. 

We use vanilla SGD and SAM \cite{foret-2020-SAM-ICLR} as baselines.
To comprehensively evaluate the performance, we have also included ASAM \cite{kwon-2021-asam-ICML}, FisherSAM \cite{kim2022fisher}, F-SAM\cite{li2024friendly}, {SGD-SALR\cite{yue2023salr}}, {Unified VaSSO\cite{oikonomousharpness}}, {SSAM-F and SSAM-D\cite{SpaSAM}} for comparison. 
These methods are the follow-up works of SAM that aim to enhance generalization. For ASAM, FisherSAM and F-SAM, we report the results in \cite{li2024friendly}. It is worth noting that these three methods were run 300 epochs for PyramidNet-110, which may lead to better results, we denote them as ASAM{\tiny(300)}, FisherSAM{\tiny(300)} and F-SAM{\tiny(300)}, respectively. {For SGD-SALR and Unified VaSSO, we refer to the results presented in \cite{yue2023salr} and \cite{oikonomousharpness}, respectively. For SSAM-F and SSAM-D, we refer to the results presented in \cite{SpaSAM}}.
Additionally, we compared several efficient SAM variants, such as LookSAM \cite{liu-2022-looksam-CVPR}, ESAM \cite{du-2022-ESAM-ICLR} and SAF \cite{du-2022-saf-NIPS}. We implemented LookSAM ourselves, while the results for ESAM, SAF are reported as presented in their original publications.
For SGD, SAM and BSAM, we conducted five independent runs with different seeds and reported the average accuracy and standard deviation. 

{To evaluate BSAM's performance on the large-scale benchmark, we conducted experiments on the ImageNet \cite{deng2009imagenet} dataset, which consists of $1,000$ classes, around $1.28$ million training images, and $50,000$ validation images, with an average resolution of $469 \times 387$. Unlike CIFAR-10/100, ImageNet is better suited to reflect real-world scenarios and offers a more robust evaluation of a method’s effectiveness. We resize the images to $224 \times 224$ for training ResNet-50 over 90 epochs, using the optimal hyperparameters for SGD, SAM, and BSAM as recommended by \cite{du-2022-ESAM-ICLR}. We conducted these experiments on ImageNet using an NVIDIA GeForce RTX 4090 GPU and compared the training speeds of BSAM and other methods. }

\begin{table}[]
\centering
\begin{tabular}{lcc}
\hline
\bf{ResNet-18}        & \bf{CIFAR-10}  & \bf{CIFAR-100} \\ \hline\hline
SGD              & 96.18\scriptsize{$\pm$0.09} & 79.89\scriptsize{$\pm$0.38} \\
SAM              & 96.74\scriptsize{$\pm$0.05} & 81.08\scriptsize{$\pm$0.27} \\ \hline
ASAM             & 96.63\scriptsize{$\pm$0.15} & {81.68\scriptsize{$\pm$0.12}} \\
FisherSAM        & 96.72\scriptsize{$\pm$0.03} & 80.99\scriptsize{$\pm$0.13} \\
F-SAM            & 96.75\scriptsize{$\pm$0.09} & 81.29\scriptsize{$\pm$0.12} \\
SGD-SALR            & / & \bf{81.73\scriptsize{$\pm$0.10}} \\
{Unified VaSSO}            & {96.34\scriptsize{$\pm$0.01}} & {80.01\scriptsize{$\pm$0.07}} \\
{SSAM-F}            & {96.81\scriptsize{$\pm$0.02}} & {81.24\scriptsize{$\pm$0.21}} \\
{SSAM-D}            & \bf{{96.87\scriptsize{$\pm$0.04}}} & {80.58\scriptsize{$\pm$0.44}} \\
\hline
LookSAM          & 96.47\scriptsize{$\pm$0.13} & 80.48\scriptsize{$\pm$0.24} \\
ESAM             & 96.56\scriptsize{$\pm$0.08} & 80.41\scriptsize{$\pm$0.10} \\
SAF              & 96.37\scriptsize{$\pm$0.02} & 80.06\scriptsize{$\pm$0.05} \\
\hline
\bf{BSAM}             & 96.82\scriptsize{$\pm$0.12} & 81.48\scriptsize{$\pm$0.18} \\ \hline \\
\bf{WideResNet-28-10} & \bf{CIFAR-10}  & \bf{CIFAR-100} \\ \hline\hline
SGD              & 96.93\scriptsize{$\pm$0.05} & 82.56\scriptsize{$\pm$0.27} \\
SAM              & 97.52\scriptsize{$\pm$0.05} & 84.71\scriptsize{$\pm$0.21} \\ \hline
ASAM             & {97.63\scriptsize{$\pm$0.13}} & 84.99\scriptsize{$\pm$0.22} \\
FisherSAM        & 97.46\scriptsize{$\pm$0.18} & 84.91\scriptsize{$\pm$0.07} \\
F-SAM            & 97.53\scriptsize{$\pm$0.11} & 85.16\scriptsize{$\pm$0.07} \\
{SGD-SALR}            & / & 83.12\scriptsize{$\pm$0.09} \\
{Unified VaSSO}            & {97.06\scriptsize{$\pm$0.09}} & {83.66\scriptsize{$\pm$0.19}} \\
{SSAM-F}            & {97.71\scriptsize{$\pm$0.23}} & {85.16\scriptsize{$\pm$0.96}} \\
{SSAM-D}            & \bf{{97.72\scriptsize{$\pm$0.24}}} & {84.99\scriptsize{$\pm$0.79}} \\
\hline
LookSAM          & 97.13\scriptsize{$\pm$0.04} & 83.52\scriptsize{$\pm$0.09} \\
ESAM             & 97.29\scriptsize{$\pm$0.11} & 84.51\scriptsize{$\pm$0.01} \\
SAF              & 97.08\scriptsize{$\pm$0.15} & 83.81\scriptsize{$\pm$0.04} \\ \hline
\bf{BSAM}             & 97.56\scriptsize{$\pm$0.07} & \bf{85.51\scriptsize{$\pm$0.14}} \\ \hline \\
\bf{PyramidNet-110}   & \bf{CIFAR-10}  & \bf{CIFAR-100} \\ \hline\hline
SGD              & 97.10\scriptsize{$\pm$0.08} & 83.38\scriptsize{$\pm$0.21} \\
SAM              & 97.65\scriptsize{$\pm$0.06} & 86.06\scriptsize{$\pm$0.16} \\ \hline
ASAM\tiny{(300)}             & 97.82\scriptsize{$\pm$0.07}         & 86.47\scriptsize{$\pm$0.09}         \\
FisherSAM\tiny{(300)}        & 97.64\scriptsize{$\pm$0.09}         & 86.53\scriptsize{$\pm$0.07}         \\
F-SAM\tiny{(300)}            & 97.84\scriptsize{$\pm$0.05}         & \bf{86.70\scriptsize{$\pm$0.14}}         \\ 
{SGD-SALR}            & / & 85.69\scriptsize{$\pm$0.10} \\ \hline
LookSAM          & 97.22\scriptsize{$\pm$0.05} & 83.76\scriptsize{$\pm$0.45} \\
ESAM             & 97.81\scriptsize{$\pm$0.01} & 85.56\scriptsize{$\pm$0.05} \\
SAF              & 97.34\scriptsize{$\pm$0.06} & 84.71\scriptsize{$\pm$0.01} \\ \hline
\bf{BSAM}             & \bf{97.96\scriptsize{$\pm$0.10}} & 86.20\scriptsize{$\pm$0.06} \\ \hline
\end{tabular}
\caption{Test accuracy (\%) comparison of various networks on CIFAR-10 and CIFAR-100.}
\label{result_cifar}
\end{table}

\subsubsection{Results}
In Table \ref{result_cifar}, BSAM outperforms both SGD and SAM on the test datasets. On the CIFAR-10 dataset, BSAM achieves the highest improvement on PyramidNet-110, with increases of 0.86\% and 0.31\% over SGD and SAM, respectively. On the CIFAR-100 dataset, BSAM delivers the most significant improvement on WideResNet-28-10, with increases of 2.95\% and 0.8\% over SGD and SAM, respectively. This indicates that BSAM is suitable for various networks and datasets, but its ability to improve generalization performance varies across different networks and datasets.

In comparison to other SAM variants designed to enhance generalization, BSAM is more effective in most cases because it considers perturbations in the gradient descent direction. Additionally, BSAM significantly outperforms efficient SAM methods, which often sacrifice model generalization performance for reduced computational costs. {Although SSAM-D outperforms BSAM on CIFAR-10 (achieving 96.87\%), it performs worse on CIFAR-100 with 80.58\%, while BSAM achieves 81.48\%. This indicates that BSAM is better at adapting to more complex datasets.} 

In general, BSAM achieved the best results on the test set in most cases on CIFAR-10 and CIFAR-100, indicating that BSAM enhances the model's generalization performance. This may be because BSAM considers both the gradient ascent and descent directions from the current parameter during training, which helps smooth the gradient descent process and ultimately guides the model toward flatter regions of the local minima. 

{Table \ref{table_imagenet} compares the test accuracy and training time of ResNet-50 trained with SGD, SAM, SSAM-F, SSAM-D, and BSAM on the ImageNet dataset. BSAM achieves the highest accuracy at 77.36\%, outperforming both SGD and SAM. {SSAM-F and SSAM-D also achieve competitive accuracies of 77.31\% and 77.25\%, but BSAM still leads, demonstrating that its optimization strategy is more effective in achieving better model performance.} However, BSAM incurs a higher training time, approximately 1.5 times that of SAM (148.1 hours  vs. 100.2 hours). This increase in training time aligns with the time complexity analysis in Section \ref{sec:bsam}. In scenarios where higher accuracy is required, the increased computational cost of BSAM is justified by its superior performance.}

\begin{figure*}[]
\centering
\subfloat[SGD]{\includegraphics[width=2.0in]{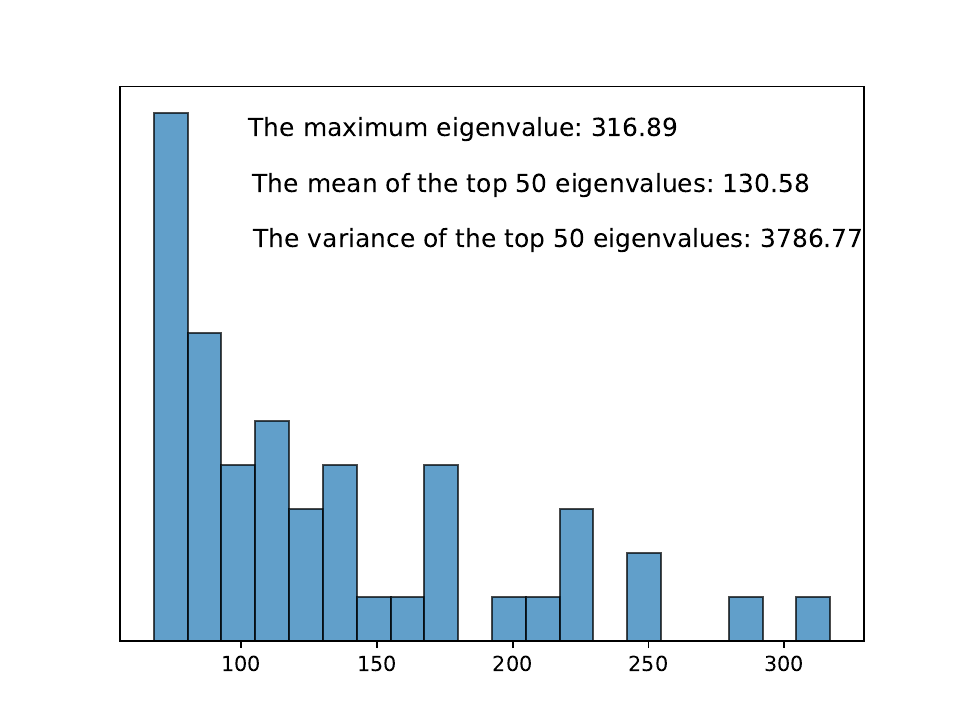}%
\label{fig:eigen_sgd}}
\hfil
\subfloat[SAM]{\includegraphics[width=2.0in]{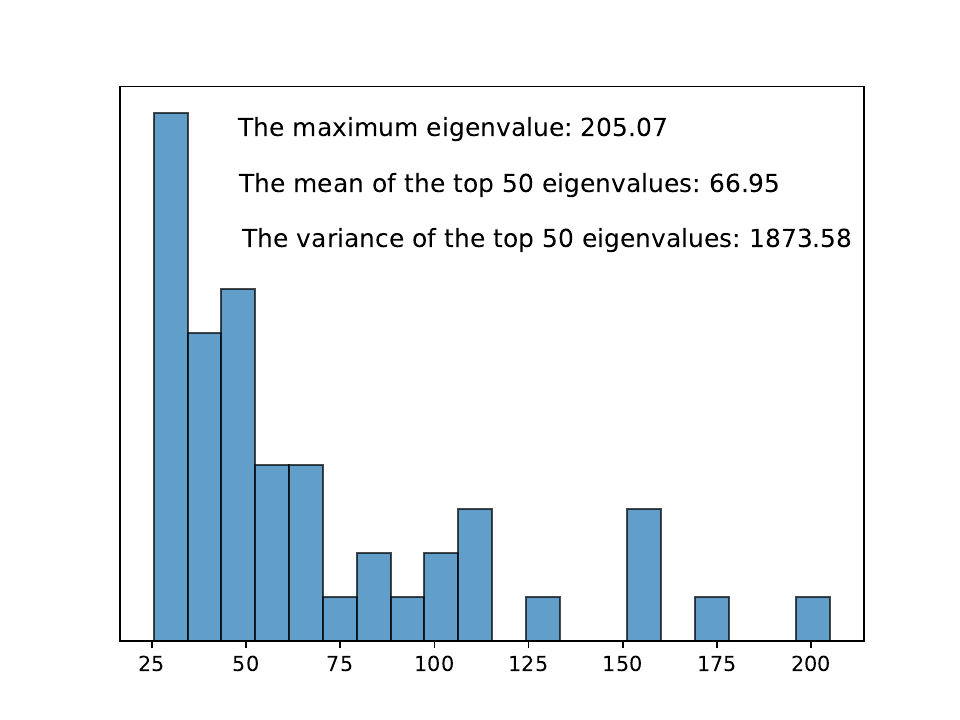}%
\label{fig:eigen_sam}}
\hfil
\subfloat[BSAM]{\includegraphics[width=2.0in]{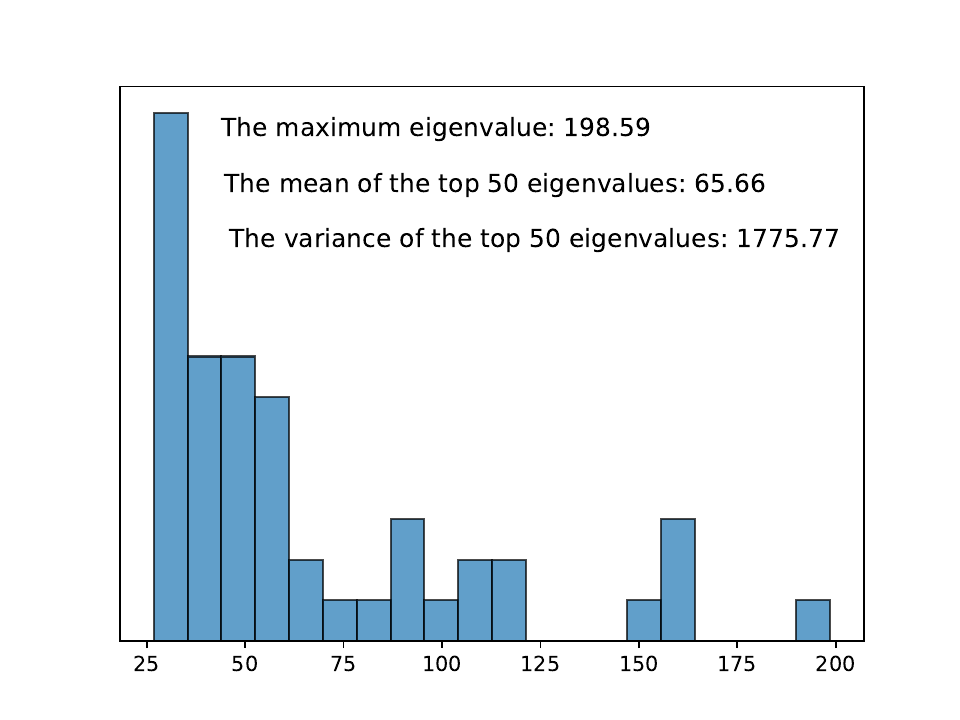}%
\label{fig:eigen_bsam}}
\caption{ The distribution of top-50 eigenvalues of Hessian on the test set of CIFAR-100 with SGD, SAM and BSAM.}
\label{fig:eigen}
\end{figure*}

\begin{table}[t]
\center
\begin{tabular}{lcc}
\hline
Methods    & Accuracy & Training Time  \\ \hline
SGD        & 76.00           &  63.5          \\
SAM        & 76.70          &  100.2         \\
{SSAM-F}        & {77.31}          &   /        \\
{SSAM-D}        & {77.25}          &   /        \\
BSAM       & \bf{77.36}      &   148.1        \\ \hline
\end{tabular}
\caption{Test accuracy (\%) and training time (hour) of ResNet-50 with SGD, SAM and BSAM on the ImageNet.}
\label{table_imagenet}
\end{table}

\subsection{Top Eigenvalues of Hessian}
{The maximum Hessian eigenvalue is commonly viewed as a potential sharpness/flatness metric \cite{keskar-2016-large_batch-ICLR} \cite{jastrzkebski2017three} \cite{chaudhari-2019-Entropy_sgd-IOP}. To offer a more quantitative assessment of sharpness compared to Fig. \ref{fig:eigen}, we compare the maximum Hessian eigenvalues obtained from SGD, SAM, and BSAM. BSAM achieves smaller maximum eigenvalues on the test set, indicating that it finds flatter minima.}

We analyze the Hessian spectra of ResNet-18 trained on CIFAR-100 with SGD, SAM and BSAM using the CIFAR-100 test set. We employ Hutchinson’s method \cite{yao2020pyhessian} to compute the eigenvalues. For each method, we report the maximum eigenvalue of Hessian, a histogram of the top-50 Hessian eigenvalues, the mean and variance of these top-50 Hessian eigenvalues. As shown in Fig. \ref{fig:eigen}, the distribution of the top-50 eigenvalues is significantly smaller for networks trained with SAM and BSAM compared to those trained with SGD. Specifically, the maximum eigenvalue for networks trained with BSAM is notably smaller than that for SGD, indicating that BSAM finds flatter minima and improves generalization. BSAM also reduces the maximum eigenvalue compared to SAM. The mean and variance of the top-50 Hessian eigenvalues for the model trained with BSAM are both smaller than those for the models trained with SGD and SAM.

\begin{table}[]
\centering
\begin{tabular}{lccccc}
\hline
\multirow{2}{*}{Datasets}  & \multirow{2}{*}{Methods} & \multicolumn{4}{c}{Noise rate} \\
                           &                          & 20\%   & 40\%  & 60\%  & 80\%  \\ \hline
\multirow{3}{*}{CIFAR-10}  & SGD                      &85.94        &70.25       &48.50       &28.91       \\
                           & SAM                      &90.44        &79.71       &67.22       &76.94       \\
                           & BSAM                     &{93.34}        & {85.58}       & {76.37}       &{78.27}       \\
                                                              \hline
\multirow{3}{*}{CIFAR-100} & SGD                      &65.42        &48.91       &31.46       &12.32       \\
                           & SAM                      &69.10        &55.06       & 35.59      & 9.82      \\
                           & BSAM                     & {71.65}       & {57.61}      & {38.71}  &  {32.69}   \\
                         
                                \hline
\end{tabular}
\caption{Results of  ResNet-18 on CIFAR-10 and CIFAR-100 with different rate of label noises.}
\label{noisy_label}
\end{table}

\begin{figure}[]
\centering
\subfloat[$\rho=0.05$]{\includegraphics[width=1.5in]{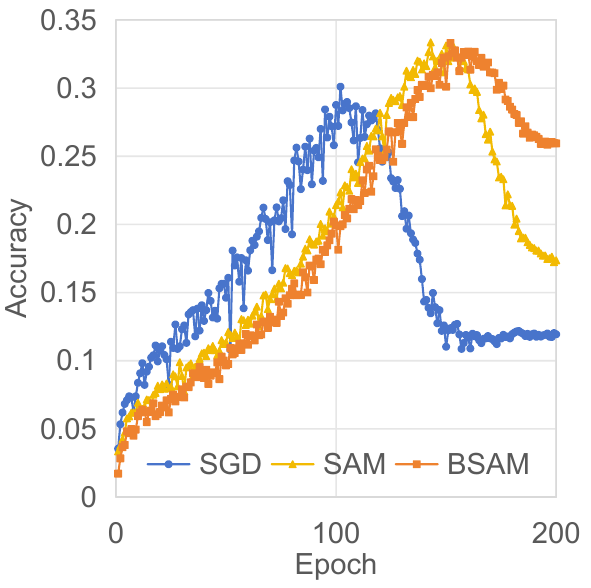}%
\label{fig:noise_sgd}}
\hfil
\subfloat[$\rho=0.1$]{\includegraphics[width=1.5in]{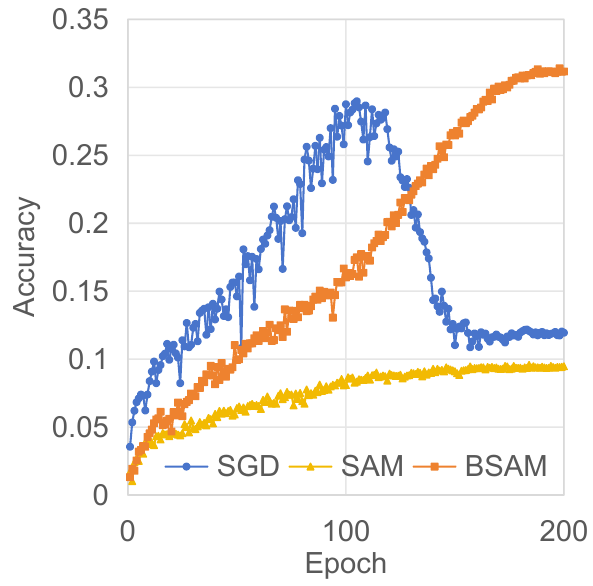}%
\label{fig:noise_sam}}
\caption{Test accuracy of optimizing ResNet-18 on CIFAR-100 using SGD, SAM and BSAM when $\rho_{max}$ is 0.05 and 0.1.}
\label{fig:noise}
\end{figure}

\subsection{Robustness to Label Noise}
Since prior studies have demonstrated that SAM is robust to label noise, this subsection evaluates the impact of applying BSAM in the classical noisy-label setting for CIFAR-10 and CIFAR-100. We evaluate the performance of BSAM under symmetric label noise by random flipping \cite{huang2019o2u}. The training settings are the same as in Section \ref{classification_setup}.

As shown in Tab. \ref{noisy_label}, for CIFAR-10 dataset, we find that the accuracy of models optimized with SGD decreases rapidly with the increase in label noise rate, dropping to just 28.91\% when the noise rate reaches 80\%. In contrast, the accuracy of models optimized with SAM and BSAM decreases more slowly with increasing label noise, maintaining over 70\% accuracy even when the noise rate reaches 80\%. For CIFAR-100, we observe a similar trend, but when the noise rate reaches 80\%, the accuracy of models trained with SAM is even lower than that of SGD. {Fig. \ref{fig:noise} shows the test accuracy of optimizing ResNet-18 on CIFAR-100 using SGD, SAM and BSAM when $\rho$ is 0.05 and 0.1. When $\rho_{max}$ is 0.1, SAM has difficulty converging stably when there are a large number of incorrect labels, as mentioned in \cite{foret-2020-SAM-ICLR}. However, the test accuracy of the models trained with BSAM steadily increase during training, indicating that BSAM can effectively converge even in the presence of high noise rate. When we set $\rho_{max}$ to 0.05, SAM is also able to converge. However, in the later stage of training, it is more prone to overfitting compared to BSAM. These observations suggest that BSAM ensures more stable convergence, facilitates reliable learning from clean labels, effectively mitigates the impact of noisy labels, and enhances generalization in high-noise environments.} Overall, BSAM consistently enhances performance compared to SGD and SAM, confirming its superior generalization capability under label noise. 

\begin{table*}[]
\centering
\begin{tabular}{lcccccccc}
\hline
Methods                        & Head   & Shoulder & Elbow  & Wrist  & Hip    & Knee   & Ankle  & Mean   \\ \hline
SimCC+Adam                     & 96.828 & \bf{95.771}   & 89.654 & 84.272 & 88.679 & 85.151 & 81.436 & 89.318 \\
SimCC+SAM                      & \bf{96.965} & 95.703   & 89.637 & 84.222 & 88.402 & \bf{85.513} & \bf{81.862} & 89.388 \\
\multicolumn{1}{l}{SimCC+BSAM} & 96.794 & 95.584   & \bf{89.995} & \bf{85.317} & \bf{89.008} & 85.392 & 81.082 & \bf{89.552} \\ \hline
\end{tabular}
\caption{Results of training SimCC on MP$\mathrm{\uppercase\expandafter{\romannumeral2}}$ with Adam, SAM and BSAM.}
\label{result_hpe}
\end{table*}

\begin{table}[]
\centering
\begin{tabular}{lccc}
\hline
Efficientnet\_b0 & SGD & SAM & BSAM \\ \hline \hline
CIFAR-10         &97.32\scriptsize{$\pm$0.07}     &97.48\scriptsize{$\pm$0.07}     &\bf{97.52\scriptsize{$\pm$0.04}}     \\
CIFAR-100        & 87.05\scriptsize{$\pm$0.11}     & 87.27\scriptsize{$\pm$0.21}     & \bf{87.40\scriptsize{$\pm$0.21}}      \\
Flowers102       &77.11\scriptsize{$\pm$0.36}     &78.04\scriptsize{$\pm$0.37}     &\bf{79.02\scriptsize{$\pm$1.10}}      \\
Standford Cars       &74.47\scriptsize{$\pm$0.55}     &75.75\scriptsize{$\pm$0.41}     &\bf{77.54\scriptsize{$\pm$0.62}}      \\
OxfordIIITPet    &87.89\scriptsize{$\pm$0.43}     &88.30\scriptsize{$\pm$0.16}     &\bf{89.42\scriptsize{$\pm$0.25}}      \\ \hline 
ResNet-50        & SGD & SAM & BSAM \\  \hline\hline
CIFAR-10         &97.28\scriptsize{$\pm$0.05}     &97.78\scriptsize{$\pm$0.02}     &\bf{97.81\scriptsize{$\pm$0.06}}      \\
CIFAR-100        &86.20\scriptsize{$\pm$0.18}     &87.48\scriptsize{$\pm$0.04}     &\bf{87.51\scriptsize{$\pm$0.05}}      \\
Flowers102       &86.74\scriptsize{$\pm$0.16}     &87.53\scriptsize{$\pm$0.13}     &\bf{88.22\scriptsize{$\pm$0.20}}      \\
Standford Cars       &80.02\scriptsize{$\pm$0.24}     &83.69\scriptsize{$\pm$0.19}     &\bf{84.36\scriptsize{$\pm$0.16}}      \\
OxfordIIITPet    &93.11\scriptsize{$\pm$0.17}     &\bf{93.14\scriptsize{$\pm$0.14}}     &93.06\scriptsize{$\pm$0.15}      \\ \hline
\end{tabular}
\caption{Test accuracy (\%) results of fine-tuning on different datasets.}
\label{transfer_learning}
\end{table}

\subsection{Transfer Learning}
Transfer learning is a technique that leverages a model trained on one task to improve performance on a related task. By fine-tuning a pre-trained model or extracting its features, it can significantly reduce the training time and data requirements for the new task, enhancing model performance even with limited data. Previous studies \cite{foret-2020-SAM-ICLR}\cite{li2024friendly} have demonstrated the effectiveness of SAM and its variants in transfer learning. In this subsection, we evaluate the performance of BSAM in transfer learning tasks. 

Specifically, we apply SGD, SAM and BSAM to fine-tuning EfficientNet-b0 and ResNet-50 (pretrained on ImageNet). Weights are initialized using the values from publicly available checkpoints, except for the final classification layer, which is resized to accommodate the new number of classes and is initialized randomly. We trained two models for 30 epochs with a batch size of 128. The initial learning rate is set to 0.01 with cosine learning rate decay. Weight decay is set to 1e-5 for EfficientNet-b0 and 1e-4 for ResNet-50 as mentioned in their papers. For SAM and BSAM, we use SGD as base optimizer, $\rho$, $\rho^{max}$ and $\hat{\rho}^{min}$ are all set to 0.05. We do not use any data augmentations for Flowers102 \cite{nilsback2008automated}, Stanford Cars \cite{krause20133d}, OxfordIIITPet \cite{parkhi2012cats}. For CIFAR-10/100, we employ the same data augmentations as previous experiments. 

As seen in Tab. \ref{transfer_learning}, SAM exhibits stronger generalization capabilities in transfer learning compared to SGD on the EfficientNet-b0 and ResNet-50, and BSAM further improves SAM. BSAM achieves optimal results in transfer learning tasks across multiple datasets for both models. Notably, it shows significant improvements on smaller datasets such as Flowers102, Stanford Cars. 

\subsection{Application to Human Pose Estimation}
2D Human Pose Estimation (2D HPE) is a computer vision task that involves detecting and analyzing human poses from images or videos. We apply SAM and BSAM to 2D HPE to evaluate their general applicability, using the SimCC method \cite{li2022simcc} for validation. The key idea of SimCC is to treat human pose estimation as two classification tasks, one for vertical and one for horizontal coordinates. We use SGD, SAM and BSAM as optimizers for the SimCC, and conduct experiments on the MP$\mathrm{\uppercase\expandafter{\romannumeral2}}$ dataset \cite{andriluka20142d}. The head-normalized probability of correct keypoint (PCKh) \cite{andriluka20142d} score is used for model evaluation. All experiments are conducted with the input image size $256\times 256$, and the batch size are set to 64.

The results are shown in Tab.~\ref{result_hpe}, reveal that BSAM performs better on key body joints such as Elbow, Wrist and Hip, while showing a decline in performance on other joints. Overall, BSAM achieves better average results compared to SGD and SAM. It's worth noting that both SAM and BSAM use the Adam optimizer as their base optimizer in these experiments. This indicates that BSAM remains effective even when Adam is used as the base optimizer.

\begin{table}[t]
\centering
\begin{tabular}{lcc}
\toprule
Methods     & {PSPNet}  & {DeepLabv3+} \\ \midrule
SGD    & 78.70\scriptsize{$\pm$0.28}      & 77.12\scriptsize{$\pm$0.28}    \\
SAM    & {79.10}\scriptsize{$\pm$0.14}    &{77.25}\scriptsize{$\pm$0.16}     \\
BSAM   & \bf{79.20}\scriptsize{$\pm$0.14}    & \bf{77.96}\scriptsize{$\pm$0.39}  \\ \bottomrule
\end{tabular}%
\caption{Mean IoU (\%) of semantic segmentation with SGD, SAM and BSAM on the VOC 2012.}
\label{table_ss}
\end{table}

\begin{table}[t]
\center
\begin{tabular}{lcc}
\hline
Methods    & ResNet-18 & MobileNetV1 \\ \hline
Full prec. & 88.72           & 85.81            \\
SGD        & 88.86\scriptsize{$\pm$0.18}           & 84.04\scriptsize{$\pm$0.13}           \\
SAM        & 89.75\scriptsize{$\pm$0.21}          & 84.72\scriptsize{$\pm$0.11}           \\
BSAM       & \bf{89.84\scriptsize{$\pm$0.11}}          & \bf{84.76\scriptsize{$\pm$0.10}}           \\ \hline
\end{tabular}
\caption{Test accuracy (\%) results of QAT with SGD, SAM and BSAM on the Cifar-10.}
\label{table_QAT}
\end{table}

\subsection{Application to Semantic Segmentation}
{
We conduct semantic segmentation experiments on the PASCAL VOC 2012 dataset \cite{everingham2010pascal}, which includes 20 object categories and one background class. Using PSPNet \cite{zhao2017pyramid} and DeepLabv3+ \cite{chen2018encoder} as representative models, we follow \cite{zhao2017pyramid}'s procedure with augmented data from \cite{hariharan2011semantic}, resulting in 10,582 training, 1,449 validation, and 1,456 test images. We compare our proposed BSAM with SGD and SAM.}

{Tab. \ref{table_ss} presents the results of semantic segmentation on the VOC 2012 dataset using different optimization methods. The results demonstrate BSAM's consistent superiority across both architectures. It shows particular strength on DeepLabv3+, surpassing SAM by 0.99\%. These consistent improvements demonstrate BSAM's superior generalization across different architectures.
}

\subsection{Application to Quantization-Aware Training}
Neural network quantization is a technique that reduces the computational and storage requirements of neural networks by approximating their weights and activations with lower precision numbers. This makes it easier to deploy models on resource-constrained devices such as mobile phones and embedded systems~\cite{jacob-2018-quantization}\cite{esser2020learned}\cite{wei2021qdrop}\cite{nagel2022overcoming}. 

We employ BSAM as the optimizer for Quantization-Aware Training (QAT) \cite{jacob-2018-quantization} to demonstrate BSAM's broader applicability. We applied SGD, SAM, and BSAM algorithms to quantize the parameters of the ResNet-18 and MobileNetV1 models to W4A4 on the CIFAR-10 dataset, the results are presented in Tab.~\ref{table_QAT}. We found that on ResNet-18, the quantized models trained with SAM and BSAM achieved results that were higher than those of the models trained with SGD, even surpassing the accuracy of the full-precision model. On MobileNetV1, SAM and BSAM also achieved better results than SGD. Overall, both SAM and BSAM enhance the generalization performance of quantized models, with BSAM achieving marginally better results than SAM.

\section{Conclusions}
{
In this paper, we address the limitations of Sharpness-Aware Minimization (SAM), particularly its focus solely on sharpness along the gradient ascent direction. By identifying the ``Flatness Indicator Problem'' (FIP) in SAM, we propose a novel approach that integrates Min-Sharpness (MinS) alongside Max-Sharpness (MaxS) to create a more effective Bilateral Sharpness (BilS) indicator. Then, we introduce Bilateral Sharpness-Aware Minimization (BSAM). The bilateral optimization strategy, as detailed in Equation \eqref{equ:loss_bsam}, ensures the loss landscape is flattened on both sides of the current parameter point, leading to flatter minima.}
Experiments across various models and datasets show that BSAM outperforms SGD and SAM, indicating that BSAM can improve the model's generalization performance. Additionally, we apply BSAM to a variety of tasks. Experiments on label noise demonstrate that BSAM exhibits superior robustness to label noise. Experiments on transfer learning, human pose estimation, semantic segmentation and model quantization, BSAM achieved better results than both SGD and SAM, indicating that BSAM can be easily applied to a variety of tasks and has broad applicability.

In future work, we will further explore the impact of perturbation directions in SAM on the optimization process, with the goal of identifying flatter regions of the minimum to enhance generalization performance. Additionally, we will also explore strategies to improve the efficiency of BSAM, thereby increasing its practical applicability.

\section*{Acknowledgments}
The work was supported by the National Key Research and Development Program of China (Grant No. 2023YFC3306401) and National Natural Science Foundation of China 62576018. This research was also supported by Zhejiang Provincial Natural Science Foundation of China under Grant No. LD24F020007, Beijing Natural Science Foundation L244043. This work was supported in part National Natural Science Foundation of China: 62236008, 62441232 and 62521007.


\appendices
\section{Proof of Eq.~\eqref{equ:hat_min}}\label{app:min_perturbation}
\begin{proof}
From Eq.~\eqref{equ:taylor_min}, we want ${\bm{\varepsilon} ^{min}}^ \top \nabla L({\bf{w}})$ to be the minimum, which mean $(- {\bm{\varepsilon} ^{min}})^ \top \nabla L({\bf{w}})$  to be the maximum. Let $\mathbf{x} =  - {\bm{\varepsilon} ^{min}}$, $\mathbf{g} = \nabla L({\bf{w}})$, suppose $p > 1$ and $\frac{1}{p} + \frac{1}{q} = 1$, we have:
\begin{equation}
\centering
\begin{aligned}
(- {\bm{\varepsilon} ^{min}})^ \top \nabla L({\bf{w}}) &= \sum\limits_{i = 1}^n {{x_i}} {g_i} \le \sum\limits_{i = 1}^n {\left| {{x_i}{g_i}} \right|} \\  &= {\left\| {{\mathbf{x}^ \top }\mathbf{g}} \right\|_1} \le {\left\| \mathbf{x} \right\|_q}{\left\| \mathbf{g} \right\|_p} \le \rho {\left\| \mathbf{g} \right\|_p}
\end{aligned}
\end{equation}
We need to find a $\left\| {\hat{\mathbf{x}}} \right\|_q \le \rho $ such that $\sum\limits_{i = 1}^n {{x_i}} {g_i} \le \rho {\left\| \mathbf{g} \right\|_p}$. 

Let $\mathbf{x} = sign(\mathbf{g}){\left| \mathbf{g} \right|^{p - 1}}$, we have:
\begin{equation}
\centering
\begin{aligned}
\sum\limits_{i = 1}^n {{x_i}} {g_i} = \sum\limits_{i = 1}^n {sign({g_i}){{\left| {{g_i}} \right|}^{p - 1}}} {g_i} = \sum\limits_{i = 1}^n {{{\left| {{g_i}} \right|}^p}}  = \left\| \mathbf{g} \right\|_p^p
\end{aligned}
\end{equation}
For $\hat{\mathbf{x}} = \rho \frac{\mathbf{x}}{{{{\left\| \mathbf{x} \right\|}_q}}}$, $\hat{\mathbf{x}}$ satisfies the condition $\left\| {\hat{\mathbf{x}}} \right\|_q \le \rho $. And the following holds:
\begin{equation}
\centering
\begin{aligned}
\sum\limits_{i = 1}^n {{{\hat x}_i}} {g_i} &= \sum\limits_{i = 1}^n {\rho \frac{{{x_i}}}{{{{\left\| \mathbf{x} \right\|}_q}}}} {g_i} = \frac{\rho }{{{{\left\| \mathbf{x} \right\|}_q}}}\sum\limits_{i = 1}^n {{x_i}{g_i}} \\
 &= \frac{\rho }{{{{(\left\| \mathbf{x} \right\|_q^q)}^{1/q}}}}\sum\limits_{i = 1}^n {{x_i}{g_i}}  = \frac{\rho }{{{{(\left\| \mathbf{g} \right\|_p^p)}^{1/q}}}}\sum\limits_{i = 1}^n {{x_i}{g_i}} \\
 &= \frac{\rho }{{{{(\left\| \mathbf{g} \right\|_p^p)}^{1/q}}}}\left\| \mathbf{g} \right\|_p^p = \rho \left\| \mathbf{g} \right\|_p^{p - p/q}\\
 &= \rho {\left\| \mathbf{g} \right\|_p}
\end{aligned}
\end{equation}
That is, $(- {\bm{\varepsilon} ^{min}})^ \top \nabla L({\bf{w}})$ is maximized when $\hat{\mathbf{x}} = \rho \frac{\mathbf{x}}{{{{\left\| \mathbf{x} \right\|}_q}}}=\rho \frac{{sign(\mathbf{g}){{\left| \mathbf{g} \right|}^{p - 1}}}}{{{{(\left\| \mathbf{g} \right\|_p^p)}^{1/q}}}}$, i.e., ${\bm{\varepsilon} ^{min}} = -\rho \frac{{sign(\mathbf{g}){{\left| \mathbf{g} \right|}^{p - 1}}}}{{{{(\left\| \mathbf{g} \right\|_p^p)}^{1/q}}}}$. Taking $p=q=2$, Eq.~\eqref{equ:hat_min} can be obtained.
\end{proof}

\section{Proof of Theorem~\ref{theorem1}}\label{app:convergence}
\begingroup
\begin{proof}
We use the notation and assumptions of Theorem~\ref{theorem1}. Let $L=L_{\mathbf{D}}$ be the full training loss, $\mathbf{g}_t=\nabla L(\mathbf{w}_t)$, and let $\mathcal{F}_t$ contain the training history before sampling $\mathbf{B}_t$. The batch consists of a fixed number of samples drawn uniformly with replacement, independently of $\mathcal{F}_t$. Write $\mathbb{E}_t[\cdot]=\mathbb{E}[\cdot\mid\mathcal{F}_t]$. All three mini-batch gradients use this same batch:
\begin{equation}\label{equ:proof_gradients}
\begin{aligned}
\mathbf{h}_t&=\nabla L_{\mathbf{B}_t}(\mathbf{w}_t),\\
\hat{\mathbf{h}}_t&=\nabla L_{\mathbf{B}_t}(\mathbf{w}_t+\hat{\bm{\varepsilon}}_t^{max}),\\
\check{\mathbf{h}}_t&=\nabla L_{\mathbf{B}_t}(\mathbf{w}_t+\hat{\bm{\varepsilon}}_t^{min}),
\end{aligned}
\end{equation}
where $\hat{\bm{\varepsilon}}_t^{max}=\rho_t^{max}\mathbf{h}_t/\|\mathbf{h}_t\|_2$ and $\hat{\bm{\varepsilon}}_t^{min}=-\rho_t^{min}\mathbf{h}_t/\|\mathbf{h}_t\|_2$. Define
\begin{equation}\label{equ:proof_update}
\begin{aligned}
\varsigma_t&=\frac{\|\hat{\mathbf{h}}_t\|_2}{\|\check{\mathbf{h}}_t\|_2},&
\mathbf{c}_t&=\hat{\mathbf{h}}_t-\varsigma_t\check{\mathbf{h}}_t,\\
\mathbf{d}_t&=\mathbf{h}_t+\mathbf{c}_t,&
\mathbf{w}_{t+1}&=\mathbf{w}_t-\eta\mathbf{d}_t.
\end{aligned}
\end{equation}
As required in Section~\ref{sec:convergence}, $\|\mathbf{h}_t\|_2>0$ and $\|\check{\mathbf{h}}_t\|_2>0$ almost surely. The ratio $\varsigma_t$ depends on the sampled batch and is not treated as a deterministic constant.

By conditional unbiasedness and the variance bound in Assumption~\ref{assumpion:bound_var},
\begin{equation}\label{equ:proof_gradient_moment}
\begin{aligned}
\mathbb{E}_t\|\mathbf{h}_t\|_2^2
&=\|\mathbf{g}_t\|_2^2+\mathbb{E}_t\|\mathbf{h}_t-\mathbf{g}_t\|_2^2\\
&\le\|\mathbf{g}_t\|_2^2+\sigma^2.
\end{aligned}
\end{equation}
We next bound the bilateral correction directly. The triangle and reverse triangle inequalities give
\begin{equation}\label{equ:proof_correction}
\begin{aligned}
\|\mathbf{c}_t\|_2
&=\|\hat{\mathbf{h}}_t-\check{\mathbf{h}}_t
       +(1-\varsigma_t)\check{\mathbf{h}}_t\|_2\\
&\le\|\hat{\mathbf{h}}_t-\check{\mathbf{h}}_t\|_2
       +|1-\varsigma_t|\|\check{\mathbf{h}}_t\|_2\\
&=\|\hat{\mathbf{h}}_t-\check{\mathbf{h}}_t\|_2
       +\bigl|\|\check{\mathbf{h}}_t\|_2-\|\hat{\mathbf{h}}_t\|_2\bigr|\\
&\le2\|\hat{\mathbf{h}}_t-\check{\mathbf{h}}_t\|_2.
\end{aligned}
\end{equation}
By mini-batch loss smoothness (Assumption~\ref{assumpion:smooth}) and the radius bound (Assumption~\ref{assumption:radii}),
\begin{equation}\label{equ:proof_perturbation}
\begin{aligned}
\|\hat{\mathbf{h}}_t-\check{\mathbf{h}}_t\|_2
&\le\tau\|\hat{\bm{\varepsilon}}_t^{max}
             -\hat{\bm{\varepsilon}}_t^{min}\|_2\\
&=\tau(\rho_t^{max}+\rho_t^{min})\le\tau R.
\end{aligned}
\end{equation}
Consequently, $\|\mathbf{c}_t\|_2\le2\tau R$ almost surely. This argument applies to the batch-dependent perturbed points without requiring unbiasedness of the gradients evaluated at those points. Using $\|\mathbf{x}+\mathbf{y}\|_2^2\le2\|\mathbf{x}\|_2^2+2\|\mathbf{y}\|_2^2$ and Eq.~\eqref{equ:proof_gradient_moment}, we obtain
\begin{equation}\label{equ:proof_update_moment}
\begin{aligned}
\mathbb{E}_t\|\mathbf{d}_t\|_2^2
&\le2\mathbb{E}_t\|\mathbf{h}_t\|_2^2
     +2\mathbb{E}_t\|\mathbf{c}_t\|_2^2\\
&\le2\|\mathbf{g}_t\|_2^2+2\sigma^2+8\tau^2R^2.
\end{aligned}
\end{equation}

Smoothness of the full training loss implies
\begin{equation}\label{equ:proof_smooth_descent}
L(\mathbf{w}_{t+1})\le L(\mathbf{w}_t)
-\eta\mathbf{g}_t^\top\mathbf{d}_t
+\frac{\tau\eta^2}{2}\|\mathbf{d}_t\|_2^2.
\end{equation}
Taking conditional expectations and applying the alignment condition $\mathbb{E}_t[\mathbf{g}_t^\top\mathbf{d}_t]\ge a\|\mathbf{g}_t\|_2^2$ (Assumption~\ref{assumption:alignment}) yields
\begin{equation}\label{equ:proof_conditional_descent}
\begin{aligned}
\mathbb{E}_t[L(\mathbf{w}_{t+1})]
&\le L(\mathbf{w}_t)-a\eta\|\mathbf{g}_t\|_2^2\\
&\quad+\tau\eta^2(\|\mathbf{g}_t\|_2^2+\sigma^2+4\tau^2R^2)\\
&=L(\mathbf{w}_t)-\eta(a-\tau\eta)\|\mathbf{g}_t\|_2^2\\
&\quad+\tau\eta^2(\sigma^2+4\tau^2R^2).
\end{aligned}
\end{equation}
For $\eta=1/\sqrt{T}$ and $T\ge\max\{1,4\tau^2/a^2\}$, we have $a-\tau\eta\ge a/2$. Taking total expectations in Eq.~\eqref{equ:proof_conditional_descent} therefore gives
\begin{equation}\label{equ:proof_total_descent}
\begin{aligned}
\mathbb{E}[L(\mathbf{w}_{t+1})]
&\le\mathbb{E}[L(\mathbf{w}_t)]
    -\frac{a\eta}{2}\mathbb{E}\|\mathbf{g}_t\|_2^2\\
&\quad+\tau\eta^2(\sigma^2+4\tau^2R^2).
\end{aligned}
\end{equation}
Summing over $t=0,\ldots,T-1$ telescopes the loss differences. Since $\mathbf{w}_0$ is deterministic,
\begin{equation}\label{equ:proof_telescoping}
\begin{aligned}
\frac{a\eta}{2}\sum_{t=0}^{T-1}\mathbb{E}\|\mathbf{g}_t\|_2^2
&\le L(\mathbf{w}_0)-\mathbb{E}[L(\mathbf{w}_T)]\\
&\quad+\tau T\eta^2(\sigma^2+4\tau^2R^2).
\end{aligned}
\end{equation}
Dividing by $a\eta T/2$ and substituting $\eta=1/\sqrt{T}$ proves the claimed bound:
\begin{equation}\label{equ:proof_convergence}
\begin{aligned}
\frac{1}{T}\sum_{t=0}^{T-1}\mathbb{E}\|\mathbf{g}_t\|_2^2
&\le\frac{2\bigl(L(\mathbf{w}_0)-\mathbb{E}[L(\mathbf{w}_T)]\bigr)}{a\sqrt{T}}\\
&\quad+\frac{2\tau(\sigma^2+4\tau^2R^2)}{a\sqrt{T}}.
\end{aligned}
\end{equation}
Finally, $L(\mathbf{w}_0)-\mathbb{E}[L(\mathbf{w}_T)]\le L(\mathbf{w}_0)-L_{\inf}$, and all constants are independent of $T$. Hence the average expected squared gradient norm is $O(T^{-1/2})$.
\end{proof}
\par\endgroup

\end{document}